\documentclass[accepted]{uai2023} 

\usepackage[american]{babel}

\usepackage{natbib} 
\usepackage{mathtools} 
\usepackage{booktabs} 
\usepackage{tikz} 

\usepackage{xcolor}
\usepackage[pagebackref=true,breaklinks=true,colorlinks,bookmarks=false]{hyperref}
\definecolor{deepred}{HTML}{940000}
\hypersetup{linkcolor=deepred}
\hypersetup{citecolor=[rgb]{0.4,0.15,0.95}}
\usepackage{url}       

\title{Human-in-the-Loop \textit{Mixup}}

\author[1]{Katherine M. Collins\thanks{Correspondence to: \href{mailto:kmc61@cam.ac.uk}{kmc61@cam.ac.uk}}
}\author[1,2]{Umang Bhatt}
\author[1,3]{Weiyang Liu}
\author[1]{Vihari Piratla}
\author[4]{Ilia Sucholutsky}
\author[2,5]{Bradley Love}
\author[1,2]{Adrian Weller}
\affil[1]{University of Cambridge}
\affil[2]{The Alan Turing Institute}
\affil[3]{Max Planck Institute for Intelligent Systems}
\affil[4]{Princeton University}
\affil[5]{University College London}

\begin{document}
\maketitle

\begin{abstract}
Aligning model representations to humans has been found to improve robustness and generalization. However, such methods often focus on standard observational data. Synthetic data is proliferating and powering many advances in machine learning; yet, it is not always clear whether synthetic labels are perceptually aligned to humans -- rendering it likely model representations are not human aligned. We focus on the synthetic data used in \textit{mixup}: a powerful regularizer shown to improve model robustness, generalization, and calibration. We design a comprehensive series of elicitation interfaces, which we release as \texttt{HILL MixE Suite}, and recruit 159 participants to provide perceptual judgments along with their uncertainties, over \textit{mixup} examples. We find that human perceptions do not consistently align with the labels traditionally used for synthetic points, and begin to demonstrate the applicability of these findings to potentially increase the reliability of downstream models, particularly when incorporating human uncertainty.
We release all elicited judgments in a new data hub we call \texttt{H-Mix}.
\end{abstract}

\section{Introduction}

Synthetic data is proliferating, fueled by increasingly powerful generative models, e.g.~\citep{goodfellow2014generative,dhariwal2021diffusion}. These data are not only consumed directly by people -- but, as training predictive models on synthetic data has been found to unlock tremendous advances in machine learning (ML)
\citep{SilverHuangEtAl16nature, simsSyntheticData, humanAnnotationsNeeded, jordon2022synthetic}, synthetic data is increasingly employed to train algorithms serving as engines of many applications humans may interact with. However, it is not always clear whether human perceptual judgments of synthetically-generated data match the generative process used to create them.

 \begin{figure}[t]
\centering
 \includegraphics[width=.95\linewidth]{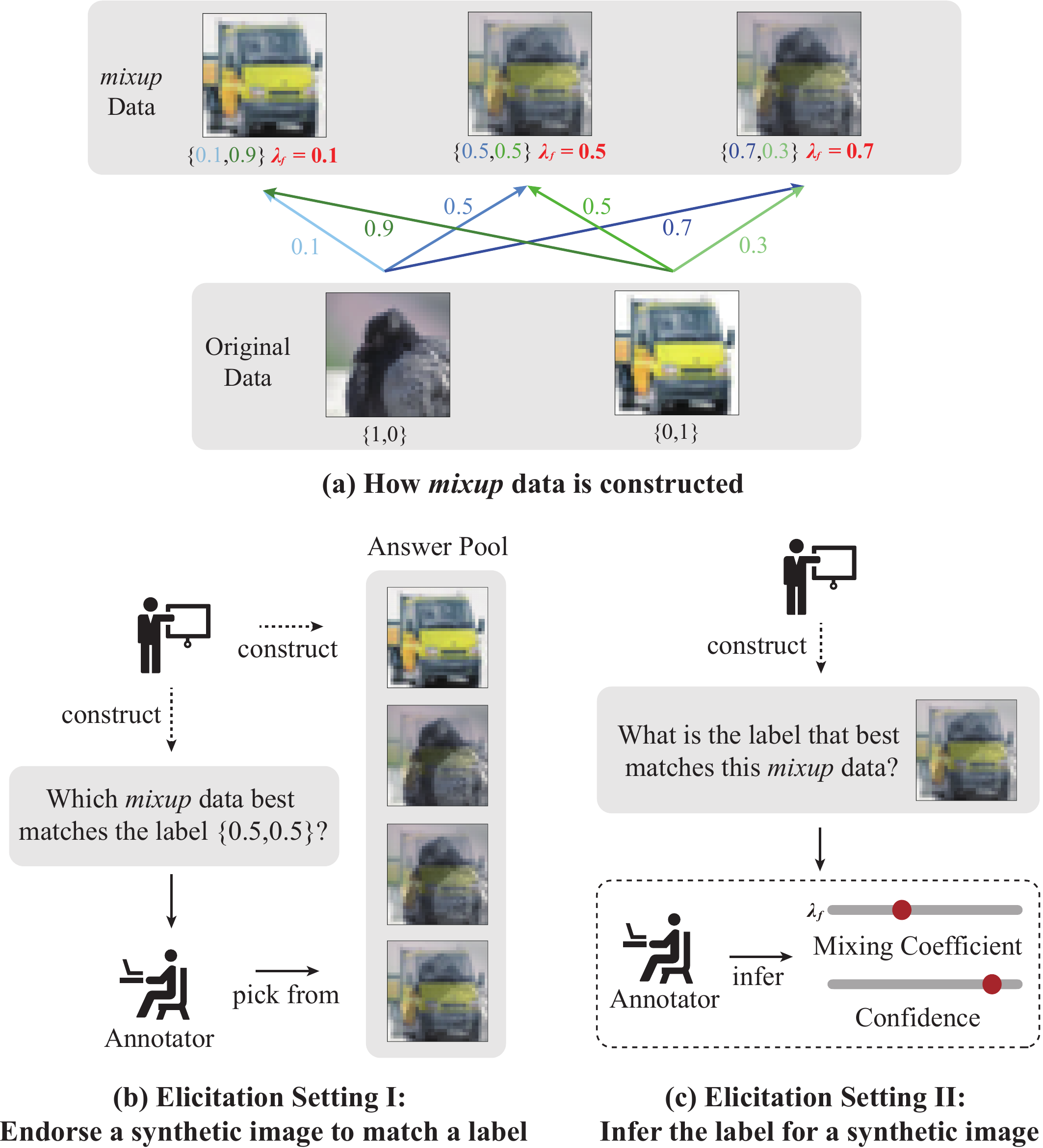}
   \caption{Framework overview. A) Synthetic data generating process used in \textit{mixup}; B) and C) depict elicitation settings: B) participants endorse a synthetic image to match a label, C) participants infer the label for a synthetic image and provide their uncertainty in the corresponding inference.}
 \label{fig:overview}
 \end{figure}

Aligning networks to match humans' perceptual inferences could be a way to further ensure model reliability, trustworthiness, downstream performance, and robustness~\citep{perceptionAlignment, chen2022perspectives, fel2022aligning, sucholutsky2023alignment}. If these data are \textit{not} aligned with human percepts, then performance potentially could be improved by altering such signals to better match the richness of human judgments: this has proven effective when aligning models with human probabilistic knowledge and perceptual uncertainty \citep{selfCiteSoftLabel, sandersambiguous, iliaSupervision}.
We argue that one ought to \textit{verify} whether synthetic data aligns with human perception, and if not, explore whether training with \textit{human-relabeled} examples improves model performance.

In this work, we take a step in this direction by focusing on \textit{mixup} \citep{mixup}: a method whereby a model is trained only on synthetic, linear combinations of conventional training examples. We 
focus on \textit{mixup} for three key reasons. First, the generative process for synthetic \textit{mixup} examples is very simple, and provides us with direct access to the ``ground truth'' generative model parameters; that is, we have precise control over the mixing coefficient used to create the mixed image. This enables us to compare any discrepancy between human perceptual judgments and this parameter explicitly. A generative model of the likes of a generative adversarial network (GAN)~\citep{goodfellow2014generative} or diffusion model~\citep{ho2020denoising} does not as easily permit these kinds of precise comparisons. Second, despite this simplicity, \textit{mixup} is a powerful and popular training-time method that has been leveraged to address model fairness~\citep{chuang2020fair}, improve model calibration \citep{onmixuptraining, zhang2022when}, and increase model robustness via regularizing the form of category boundaries learned implicitly~\citep{mixupRobustness, mixupE}. \textit{mixup} is frequently used as a strong benchmark for many new data augmentation and regularization techniques \citep{hendrycks2019augmix, hendrycks2022pixmix}. Third, prior work in human categorical perception -- revealing that humans show non-linear ``warping'' effects along category boundaries \citep{harnad2003categorical, folstein2013category, goldstone2010categorical} -- suggests that humans \textit{will} differ in their percepts from the linear category boundaries encouraged by \textit{mixup}.

To that end, we consider whether \textit{mixup} labels match human perception, and if not, how the labeling scheme can be improved to better align with human intuition -- and human uncertainty -- to potentially enhance model performance. We focus on two flavors of elicitation: 1) having participants ``construct'' a midpoint between categories by selecting from a set of synthetic images, and 2) eliciting traces of humans' broader category boundary across a range of mixed images by having participants directly intervene on the synthetic label, along with their uncertainty in their judgments. We design three online elicitation interfaces to address these questions, which we offer as The Human-in-the-Loop Mixup Elicitation Suite (\texttt{HILL MixE Suite}). We collect judgments from over 150 humans on these synthetically combined images, which we release in a dataset we call ``Human Mixup'' or \texttt{H-Mix}\footnote{All data, elicitation interfaces, and experiment code will be included in our \href{https://github.com/cambridge-mlg/hill-mixup}{repository}.}. We then demonstrate one of the possible use cases of this data: as adjusted training data for deep networks to improve model generalization, calibration, and adversarial robustness.  We depict our general framework in Fig. \ref{fig:overview}. Our data (\texttt{H-Mix}) and general elicitation paradigm (e.g., \texttt{HILL MixE Suite}) could support a range of downstream applications: from serving as new training labels for machine learning or benchmarking model alignment to auditing synthetic data, and informing cognitive science studies, among others. We see our work as a step in the exciting direction of a human-centric perspective on synthetic data powering many ML algorithms, which emphasizes the potential utility of \textit{human} uncertainty in human-in-the-loop systems. 

\section{Problem Formulation}
\subsection{Decoupling Data and Label Mixing in \textit{mixup}}

We first review \textit{mixup}~\citep{mixup} and explicate the recipe by which synthetic examples are created. We employ the nomenclature and notation around ``\textit{mixup} policies'' from \citep{automix}. 
We assume access to a finite set of $N$ samples $\{(x_1, y_1), (x_2, y_2, \cdots, (x_N, y_N)\}$.
\textit{mixup} training consists of constructing synthetic training examples $(\tilde{x}, \tilde{y})$ via linear combinations of pairs of the training observations $(x_i, y_i), (x_j, y_j)$ for $i, j \in [1, N]$, corresponding to the following data and label mixing functions:
\begin{equation}
    \textnormal{Data Mixing:}~~f(x_i, x_j, \lambda_f) = \lambda_f x_i + (1-\lambda_f) x_j = \tilde{x}
\end{equation}
\begin{equation}
    \textnormal{Label Mixing:}~~g(y_i, y_j, \lambda_g) = \lambda_g y_i + (1-\lambda_g) y_j = \tilde{y}
\end{equation}
where $\lambda_f$ and $\lambda_g$ are defined as the \textit{\textbf{data mixing coefficient}} and \textit{\textbf{label mixing coefficient}}, respectively. We refer to the combined images $x_i, x_j$ and their labels $y_i, y_j$ as the \textit{\textbf{endpoints}}. For a specified mixing coefficient $\lambda$, we denote the resultant image as $\tilde{x}$. \textit{mixup} typically assumes $\lambda_f=\lambda_g$.
We instead decouple the data and label mixing functions to permit a more general formulation where the data and label mixing functions can have different coefficients.

\subsection{Human-in-the-Loop \textit{mixup}}

Our decoupling allows us to probe whether human percepts align with either the mixing policy over the observations ($f$) or the targets ($g$). Human alignment of these mixing policies could be important for several reasons. First, we may want to understand how well the synthetic data used to power many models deployed on the web matches human perceptual judgments, thus ensuring model trustworthiness. Second, given that these policies do afford \textit{mixup} downstream niceties--such as improved generalization, robustness, and calibration-- we believe it is worth exploring whether modulating such data to be more human-aligned can yield similar, or better, performance boosts. We, therefore, pose two questions to separate groups of human participants to better elucidate alignment of the \textit{mixup} synthetic data construction: 
 
\textbf{RQ1:} What $\tilde{x}$ do participants believe matches a given $\tilde{y}$?

\textbf{RQ2:} Conditioned on $\tilde{x}$, what do humans perceive as $\tilde{y}$? 

Unless otherwise noted, we focus on the setting where we maintain the structural form of $f$ and $g$; that is, they are each parameterized by a single mixing coefficient. We discuss alternative functional forms which may more flexibly capture the richness of human percepts of these synthetically-constructed images in the Supplement.

\section{Selecting a Matching Midpoint (RQ1)}

We first consider holding $g$ fixed and \textit{creating} a perceptually-aligned input. We liken this setting to counterfactual data creation from \citep{counterfactuallyAugmentedHuman}. 

\subsection{Problem Setting}
In our setup, we inform participants that they will observe samples combined from particular categories $y_i, y_j$. We fix the label mixing coefficient, $\lambda_g$ (here, to 0.5 -- but our procedure could be extended to arbitrary mixing coefficients) and ask participants to construct a viable $\tilde{x}$ that would be perceived as the $\lambda_g$ mixture of the categories. Ideally, we may want to see what kind of example the participant may select from the full space of possible examples (in our case, images); for simplicity, we restrict that participants choose a $\tilde{x}$ from a set of $M$ pre-constructed linear interpolations which we refer to as $\{\tilde{x}_j\}_{j=1}^{M}$, which we refer to as $\tilde{X}_M$. Each $\tilde{x}_j$ is the result of executing $f$ for a given $\lambda_f$. Here, we consider a sweep of over the mixing coefficients $[0.0, 0.1, ... 0.9, 1.0]$. 
From their selected image, we can uncover how their perception of the data-generating process differs relative to what was actually used to create said selected image.
\subsection{Elicitation Paradigm}
We design two means of eliciting people's selection of a $\tilde{x}$:

\begin{enumerate}
    \item Interface 1 (\texttt{Construct}): participants use their keyboard to iterate over $\tilde{X}_M$ (ordered), where key presses increment or decrement $j$ by one such that $\tilde{x}_j$ are cycled through at increments of $0.1$. One mixed example is displayed on the screen at a time. Participants press ``Next'' when they are happy with the selected $\tilde{x}_j$. 
    \item Interface 2 (\texttt{Select-Shuffled}): participants see all $\tilde{x} \in \tilde{X}_M$ on the screen at once. Mixed examples are \textit{shuffled} and thus presented in an unordered fashion. Participants indicate their selection by clicking on the $\tilde{x}_j$ they think best matches $\lambda_g$. 
\end{enumerate}


Example interfaces, and design rationales, are depicted in the Supplement. As mentioned, participants are explicitly told the categories being combined ($y_1, y_2$) and are asked to indicate the image that they think is most likely to be perceived as the 50/50 combination of the mixed images by \textit{100 other crowdsourced workers}. Such elicitation language is drawn from \citep{efficientElic}, following a recommended practice in high-fidelity human subject elicitation whereby participants are asked to assume a third-person perspective when responding \citep{prelec2004bayesian, oakley2010shelf}.

\paragraph{Stimuli and Participants} We focus on a random subset of the \texttt{CIFAR-10} test images, a dataset containing low-resolution images drawn from ten categories of objects and animals (e.g., truck, ship, cat, dog) \citep{cifar10}. We use the test set as this permits downstream comparisons against \texttt{CIFAR-10H}: an expansive set of approximately 51 human annotators' judgments about each example \citep{peterson2019human, battleday2020capturing}. From each unique category combination (e.g., truck-dog, ship-cat, cat-dog), we sample $6$ random images from each of the categories and linearly combine them in pixel-space.  We sample $249$ such image pairings, and for each, we sweep over the space of $11$ mixing coefficients incremented by $0.1$ between $\lambda_f = 0.0$ and $\lambda_f = 1.0$ (totaling $2739$ synthetically mixed images in total). We recruit a total of 70 participants from Prolific \citep{palan2018prolific} and hosted on Pavlovia. 45 participants were allocated to \texttt{Construct}, which was subdivided into two conditions based on the starting point of the selection: 23 participants started at the $\lambda_f = 0.9$ mixing coefficient, and 22 participants were assigned always starting at $\lambda_f = 0.1$. The remaining 25 participants were allocated to \texttt{Select-Shuffled}. Further details are included in the Supplement. 


\subsection{Investigating Data Mixing Alignment}

\begin{figure}[h!]
 \begin{center}
 \includegraphics[width=0.6\linewidth]{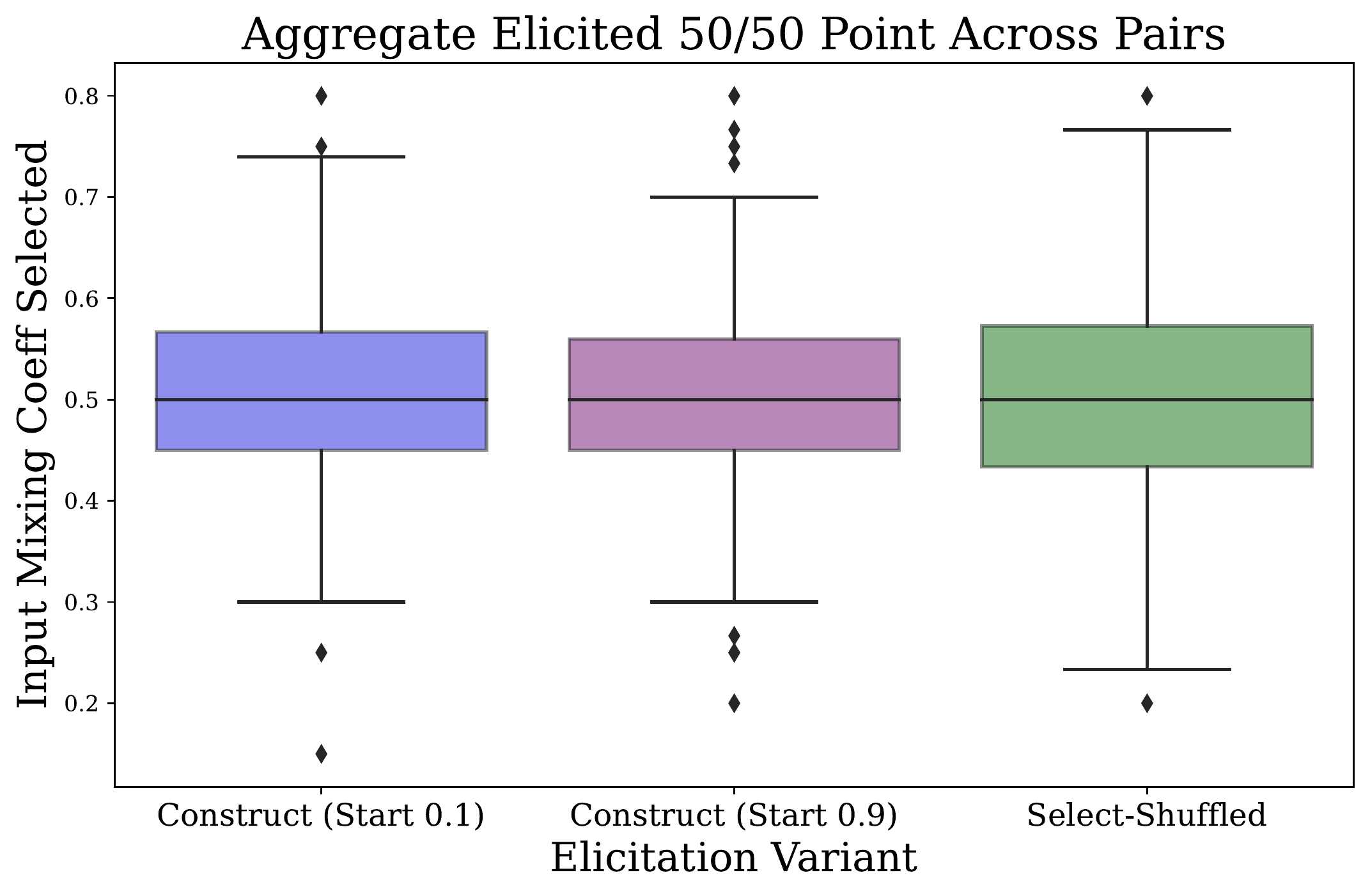}
   \caption{Averaging human participants' selections per image pair reveals the typical pair is minimally relabeled.}
 \label{fig:aggSelectRelabel5050}
 \end{center}
 \end{figure}

\begin{figure}[t!]
 \begin{center}
 \includegraphics[width=0.6\linewidth]{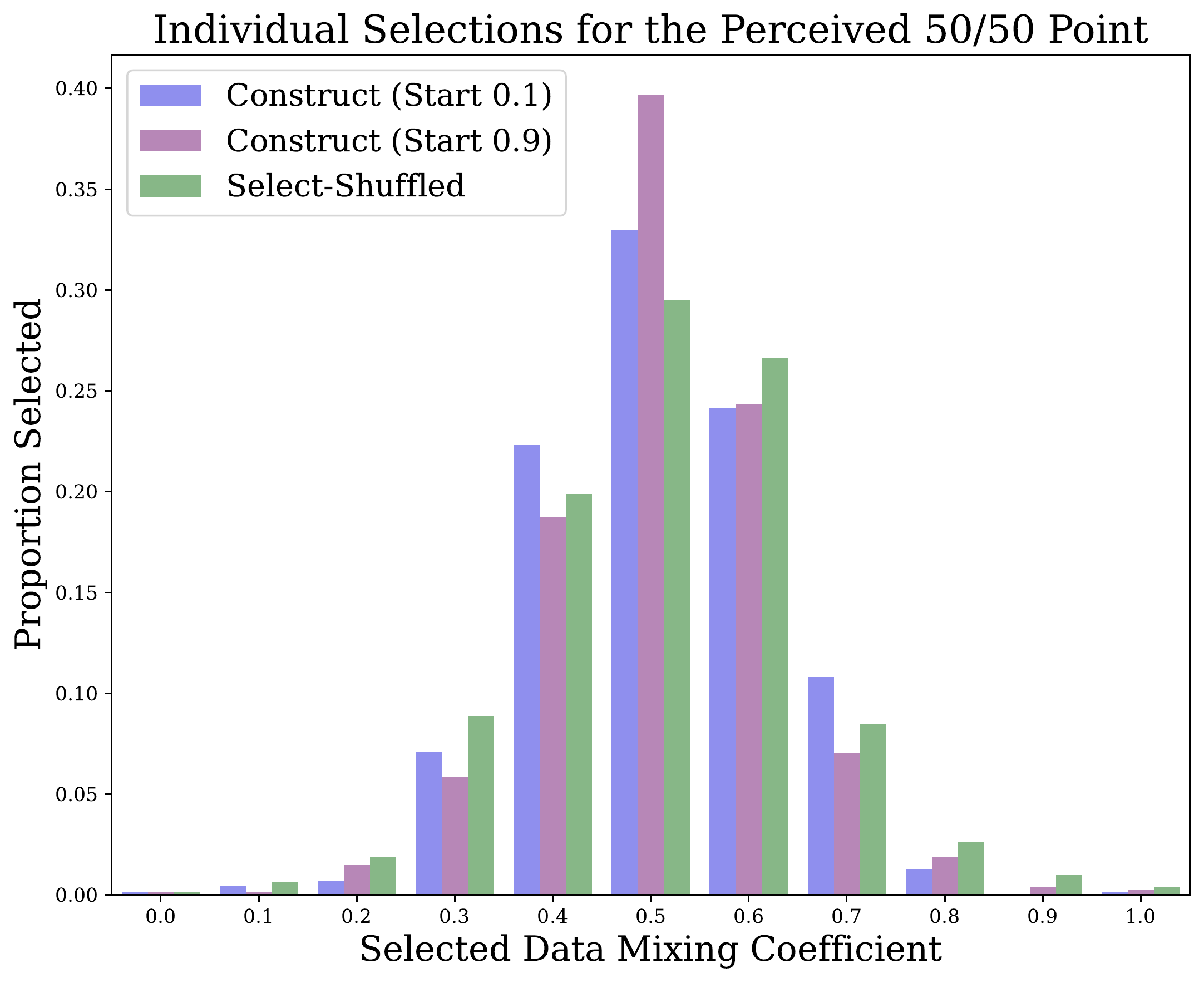}
   \caption{Participants do not always endorse the 50/50 point suggesting misalignment in the data labeling policy. The bar plot depicts extracted mixing coefficient of individuals' selections for the perceptually-aligned midpoints.}
 \label{fig:individSelect5050}
 \end{center}
 \end{figure}

\begin{figure*}[h!]
 \begin{center}
 \includegraphics[width=0.85\linewidth]{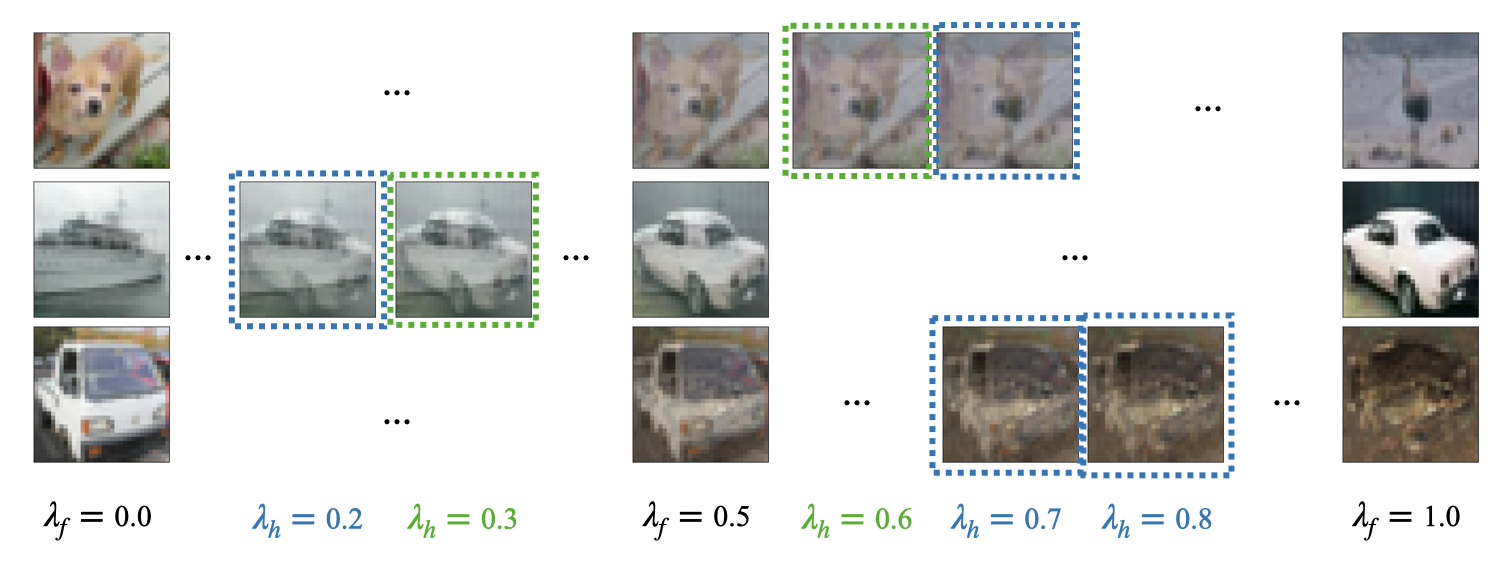}
   \caption{Example image pairs where substantial relabeling of the 50/50 point was recommended across all interface types. Synthetic images highlighted in blue received the most endorsements from participants across all interface types, with images in green receiving the second most. For row three, participants were split equally between two selections. The mixing coefficient ($\lambda_f$ or $\lambda_h$) used to construct the images is shown along the bottom.}
 \label{fig:highRelabel5050}
 \end{center}
 \end{figure*}

We find that, in aggregate, humans' selections indicate alignment with the underlying mixing coefficient (see Fig. \ref{fig:aggSelectRelabel5050}), which is stable across elicitation methodology. However, we cannot conclude from these data that the \textit{mixup} data policy is aligned with humans. If we look at the selections made by individual humans, we see that a substantial portion endorsed a $\tilde{x}$ which differed from that which would naturally be assumed in \textit{mixup} (see Fig. \ref{fig:individSelect5050}). Example image pairs that yield high relabeling across interface types are shown in Fig. \ref{fig:highRelabel5050}. We identify 9 such image pairs that are highly relabeled (which we define as $|\lambda_{h} - 0.5| \geq 0.15$, where we let $\lambda_{h}$ be the mixing coefficient used to generate the $\tilde{x}$ selected by humans) across interface types. This picture suggests that indeed human percepts are \textit{not} consistently aligned with the synthetic data construction process -- and that perhaps with a larger set of stimuli, more such examples can be recovered. Note, there are a total of 101 image pairs that are endorsed by at least one interface as in need of high relabeling. More work is needed to elucidate whether discrepancies in relabeling were induced by the varied interface design or simply individual differences among the participants recruited.

\textit{Takeaways} These data suggest that while in general, the 50/50 combined image is recoverable -- at an individual level, such percepts are more nuanced. Our data, which we include as part of \texttt{H-Mix}, indicate systematic differences in perceptions of synthetically-constructed data. These differences emerge somewhat robustly across elicitation types. We next turn to richer traces of humans' perceptual representations of these synthetically-generated data.

\section{Elucidating Alignment of the Label Mixing Policy (RQ2)}


The above elicitations have focused only on the 50/50 point; however, \textit{mixup} trains on synthetically-generated images sampled for a wide range of mixing coefficients. It, therefore, warrants study to analyze human perceptual alignment over a richer spectrum of mixing coefficients. We consider instead eliciting humans' judgments over what the label mixing coefficient $\lambda_g$ ought to be. Studying the alignment of $g$ could push forward a deeper understanding of what the data often used to train \textit{mixup} and similar methods even means to humans, and potentially further motivate the design of alternative relabeling schemes (see Section 5).  
We therefore now focus on utilizing human input to design a perceptually-aligned target \textit{mixup} policy $g_h$. 

\subsection{Problem Setting}

We assume $f$ is a linear mixing policy over inputs employed in \citep{mixup}. To form our human-aligned target policy, we want to find a function $g_h(y_i, y_j, \lambda) = \tilde{y}$ such that $\tilde{y}$ perceptually corresponds to the associated mixed input $f(x_i, x_j, \lambda) = \lambda x_i + (1-\lambda) x_j = \tilde{x}$. How do we get $\tilde{y}$ from people efficiently?

We consider matching $\lambda_g$ to what humans \textit{infer} $\lambda_f$ to be. In this setup, we assume humans are aware of the generative processes $f$ and $g_h$, and are shown the mixed image $\tilde{x}$ and underlying labels $y_i, y_j$. People are then tasked with forming a probabilistic judgment as to what the underlying mixing coefficient is that generated the observed image $\tilde{x}$ when given the underlying $y_i, y_j$ -- e.g., judging $P(\lambda_f | \tilde{x}, y_i, y_j)$. 

If human perception is aligned to the underlying linear \textit{mixup} policies, then the human predicted mixing coefficient $\lambda_h$ should be equivalent to $\lambda_f$, rendering $\lambda_f = \lambda_g = \lambda$ a sensible mixing scheme. However, if human estimates are not aligned, we may consider setting $\lambda_g = \lambda_h$ to make $g$ yield a $\tilde{y}$ which best corresponds to humans' percepts of $\tilde{x}$.

\subsection{Elicitation Paradigm}

To elicit such information, we design a new interface where subjects infer the mixing coefficient between two given labels. We show each worker a mixed image and tell them the categories that were mixed to generate the image. Participants also provide us with their \textit{uncertainty} in their inference. As some image combinations appear quite convoluted, we reason that subjects' confidence in their inference -- or lack therefore -- may provide interesting signals as to the perceptual sensibility of the mixed images. We run our relabeling experiment on $N=81$ participants again through Prolific \citep{palan2018prolific}. Further details are included in the Supplement.


\paragraph{Stimuli selection} Similar to Section 3.2, we sample images to mix from \texttt{CIFAR-10} \citep{cifar10}. We do so in a class-balanced fashion: $46$ mixed images are sampled for each of the $45$ possible class combinations, resulting in $2070$ total stimuli. Each mixed image is formed by constructed by selecting a data mixing coefficient $\lambda_f \in \{0.1, 0.25, 0.5, 0.75, 0.9\}$.


\subsection{Validating the Mixing Coefficient against Human Responses}

We now compare the human-inferred mixing coefficient against the generating coefficient and analyze participants' uncertainty in such inferences. We also conduct a preliminary exploration into the relationship between participants' predicted uncertainty and the ambiguity of the underlying images being combined.


\begin{figure}[t!]
 \begin{center}
 \includegraphics[width=0.7\linewidth]{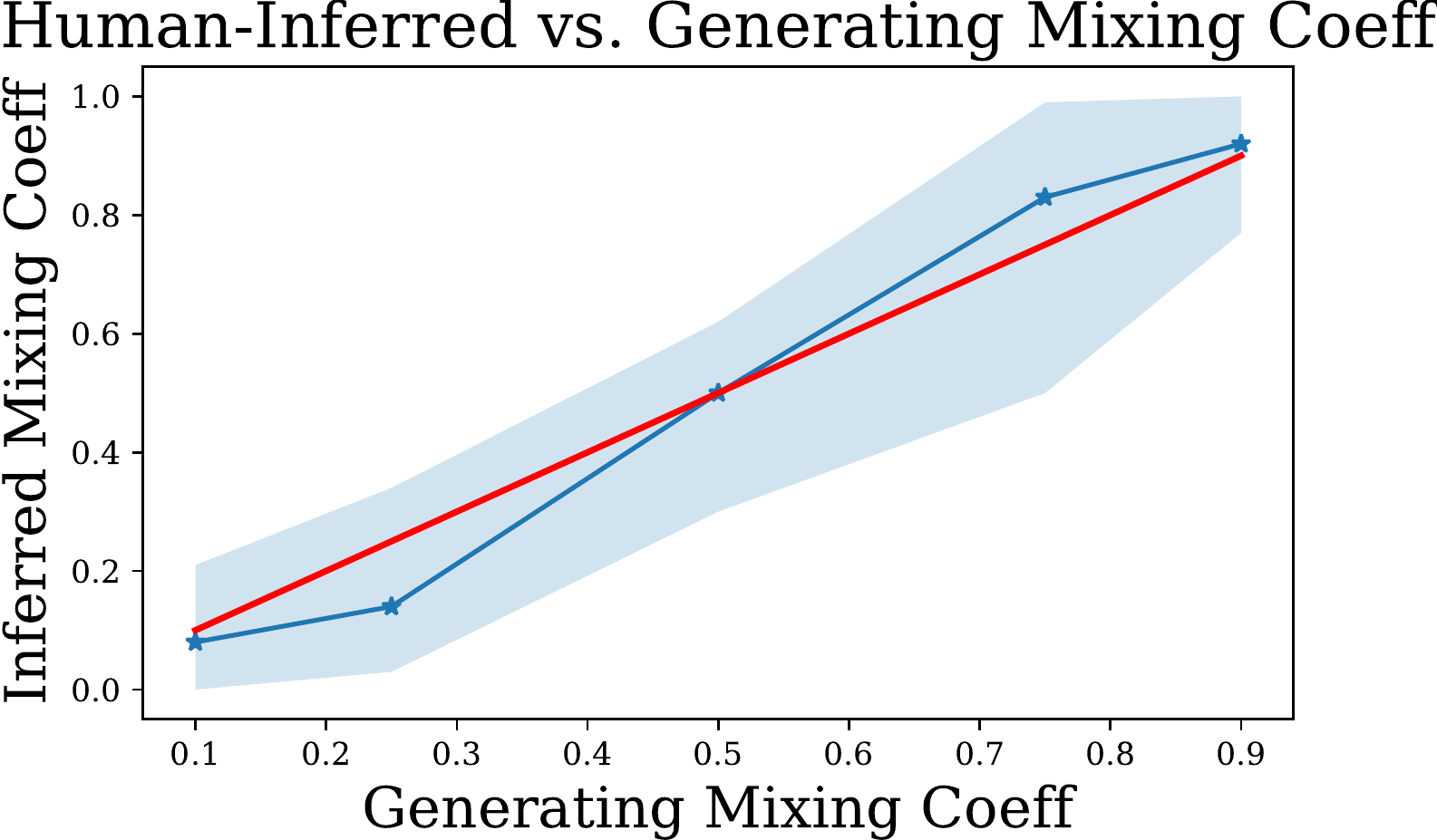}
 \caption{We uncover a sigmoidal relationship between humans' inferred mixing coefficient ($\lambda_h$, blue) as compared to the mixing coefficient used to generate the image ($\lambda_g$, red) suggestive of misalignment. We depict the median, along with the 25th and 75th percentiles. The red line indicates what the exact parallel between $\lambda_h$ and $\lambda_f$ would look like (highlighting perceived human deviation). }
 \label{fig:averageHumanRelabeling}
 \end{center}
 \end{figure}

\subsubsection{Relationship between Generating Mixing Coefficient and Alignment} 

\begin{figure}[t!]
 \begin{center}
 \includegraphics[width=0.8\linewidth]{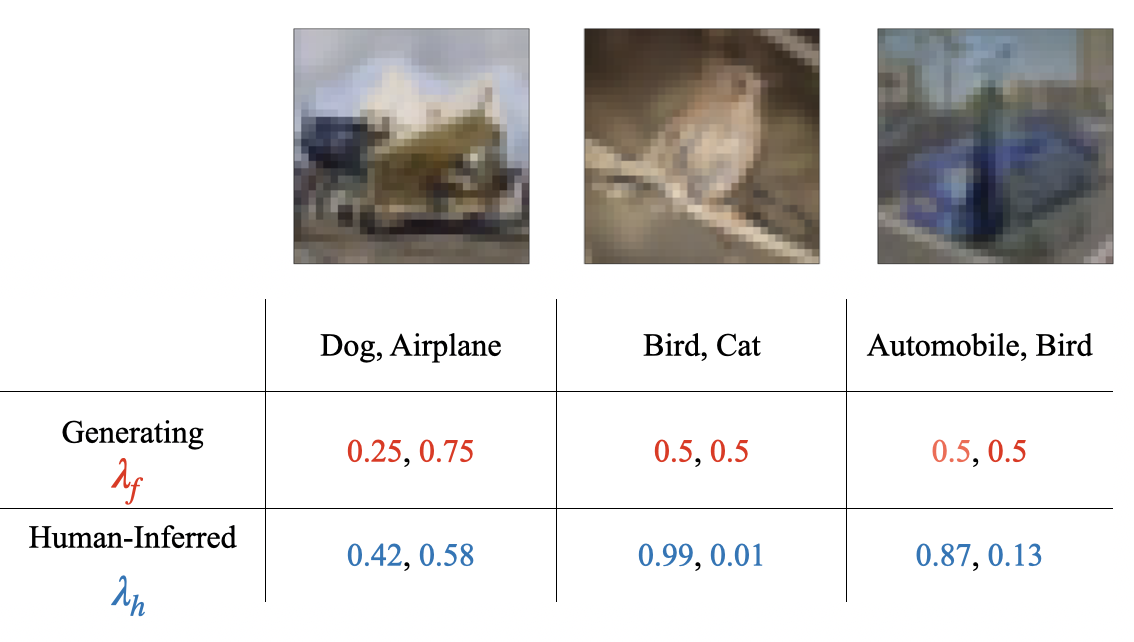}
   \caption{Examples of average human relabelings of the generating mixing coefficient reveal discrepancies.}
    \label{fig:exampleRelabel}
 \end{center}
 \end{figure}
\begin{figure*}[h!]
 \begin{center}

 \includegraphics[width=0.99\linewidth]{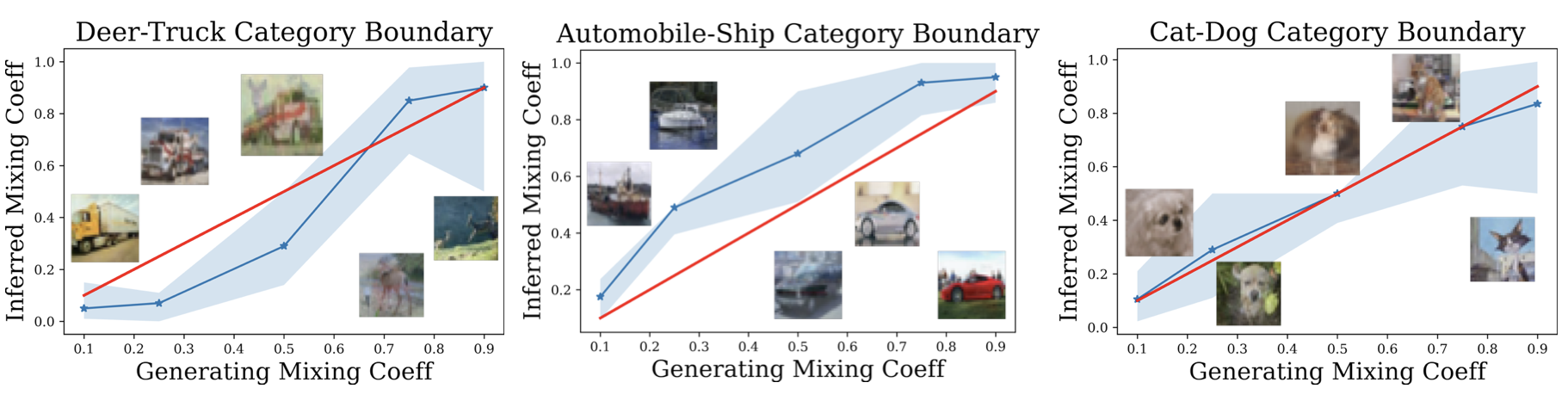}
    \caption{``Category boundaries'' elicited from humans display a diverse structure. Many -- though not all -- deviate from linearity assumed in \textit{mixup}. We overlay examples of synthesized stimuli, ordered by the $\lambda_f$ used to create them.} 
    \label{fig:categoryBoundaries}
 \end{center}
 \end{figure*}

We consider whether participants recover the data mixing coefficient: in Fig. \ref{fig:averageHumanRelabeling}, we show the median relabeling for images at given coefficients. We observe a non-linear, roughly sigmoidal structure to human relabelings, consistent with past research in human categorical perception \citep{harnad2003categorical, goldstone2010categorical, folstein2013category, morphSpacesShape}. The aggregate recovery of the 50/50 point corroborates our findings in RQ1. However, we find that the picture is nuanced: wide confidence bounds suggest there are mixed images for which inferred mixing coefficients are substantially different from the parameterization assumed in \textit{mixup}. Qualitative inspection of averaged relabelings for particular images (Fig. \ref{fig:exampleRelabel}) -- and across category pairs (Fig. \ref{fig:categoryBoundaries}) -- also reveals such misalignment. We recommend future work to investigate why particular category pairs, for this dataset, are yielding different boundaries.


 \subsubsection{Analyzing Human Uncertainty}

We next look closer at the reported human uncertainty in the mixing coefficient. First, we investigate whether human uncertainty estimates depend on the mixing coefficient assigned. Indeed, we do observe that humans’ uncertainty tracks with the mixing coefficient (see Table \ref{tab:uncertainty_per_coeff}); participants have the lowest confidence (i.e., highest uncertainty) for images generated from $\lambda_f = 0.5$. 

\begin{table}[h!]
\centering
\caption{Participants' average reported confidence, or uncertainty, in their inference of the mixing coefficient (higher confidence means less uncertainty). Error bars indicate standard deviation across participants. The mixing coefficient here is computed as $|0.5 - \lambda_f|$ due to symmetry (a mixing coefficient of $0.1$ is as extreme as $0.9$).}
\label{tab:uncertainty_per_coeff}
\begin{tabular}{@{}ll@{}}
\toprule
\textbf{Mixing Coefficient} & \textbf{Reported Confidence} \\ \midrule
0.1                    & 0.79 $\pm$ 0.17              \\
0.25                   & 0.72 $\pm$ 0.20              \\
0.5                    & 0.63 $\pm$ 0.20              \\ \bottomrule
\end{tabular}

\end{table}

Additionally, while intuitive, we probe whether there are specific predictors of when and why a mixed image may be hard to label -- e.g., perhaps images which are naturally ambiguous become even more muddled when combined. We use the entropy of the \texttt{CIFAR-10H} labels as a measure of image ``ambiguity''\citep{peterson2019human, battleday2020capturing}. Recall, \texttt{CIFAR-10H} labels are constructed from many annotator's judgments about the most probable image category; entropy is therefore computed over the frequencies of these class selections and captures some sense of the amount of disagreement between annotators. 

\begin{figure}[h!]
 \begin{center}
 \includegraphics[width=0.6\linewidth]{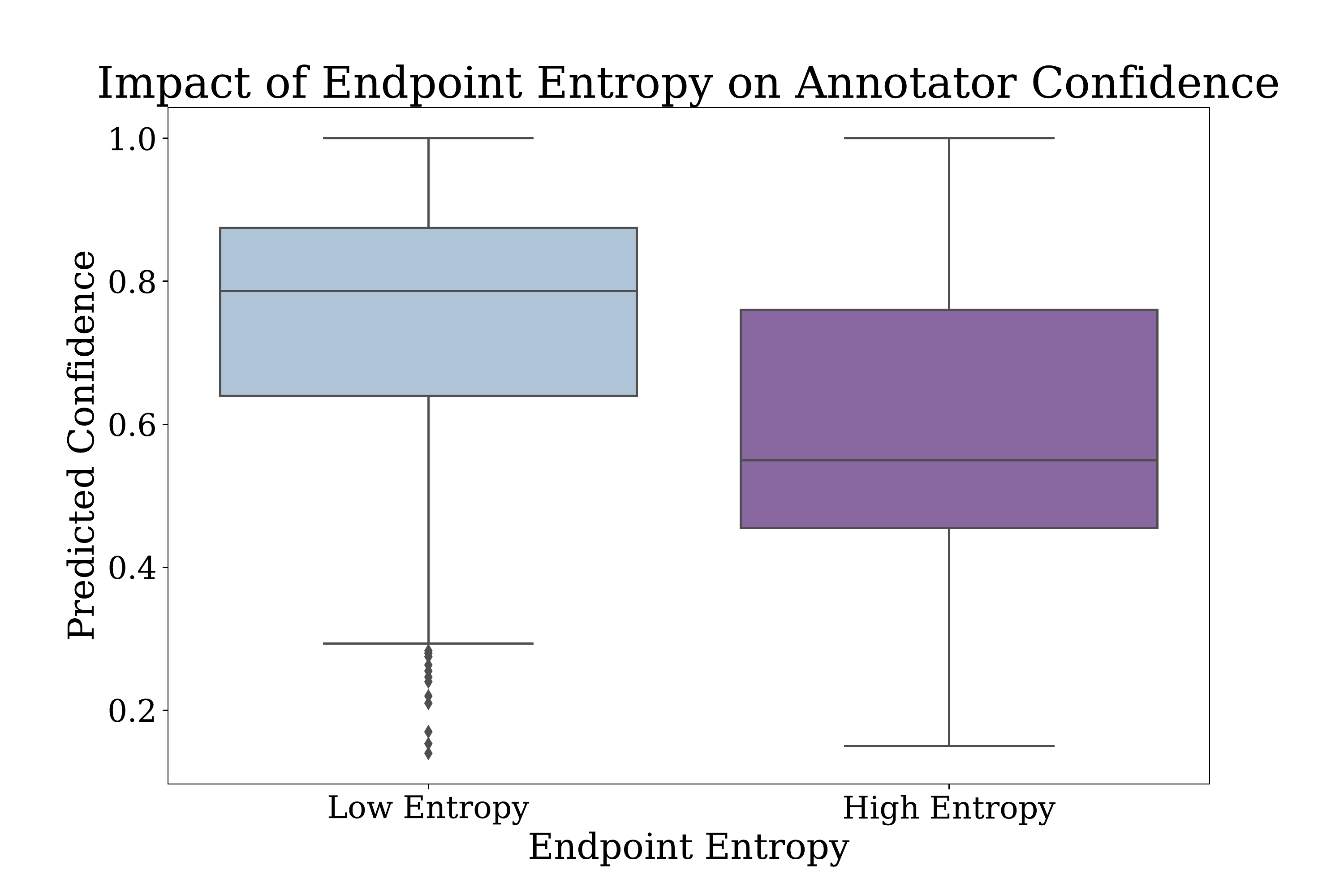}
\caption{Uncertainty reported by annotators in their inference of $\lambda$, as a factor of whether the combined labels $y_i, y_j$ are high or low entropy. Entropy is measured over the \texttt{CIFAR-10H} human-derived labels. }
 \label{fig:decompEndpoints}
 \end{center}
\end{figure}

We compare humans' elicited confidence in their mixing coefficient, and the amount of relabeling ($| \lambda_h - \lambda_f |$) against the entropy of the \texttt{CIFAR-10H} labels of the images being combined. We find in Fig. \ref{fig:decompEndpoints} that if both endpoints are high entropy under \texttt{CIFAR-10H} (where we consider ``high'' being entropy $\geq$ 0.5), participants report markedly lower confidence in their inference than if both endpoints have low entropy (entropy $\leq$ 0.1). However, we do not find a significant effect of endpoint entropy and amount of relabeling. This suggests that the ambiguity of the underlying images being mixed plays some role in determining when the resulting synthetic image may be hard to label, but there remains a question as to what can predict high amounts of relabeling from participants. We leave these questions for future investigation.



\begin{figure}[h!]
 \begin{center}
 \includegraphics[width=0.65\linewidth]{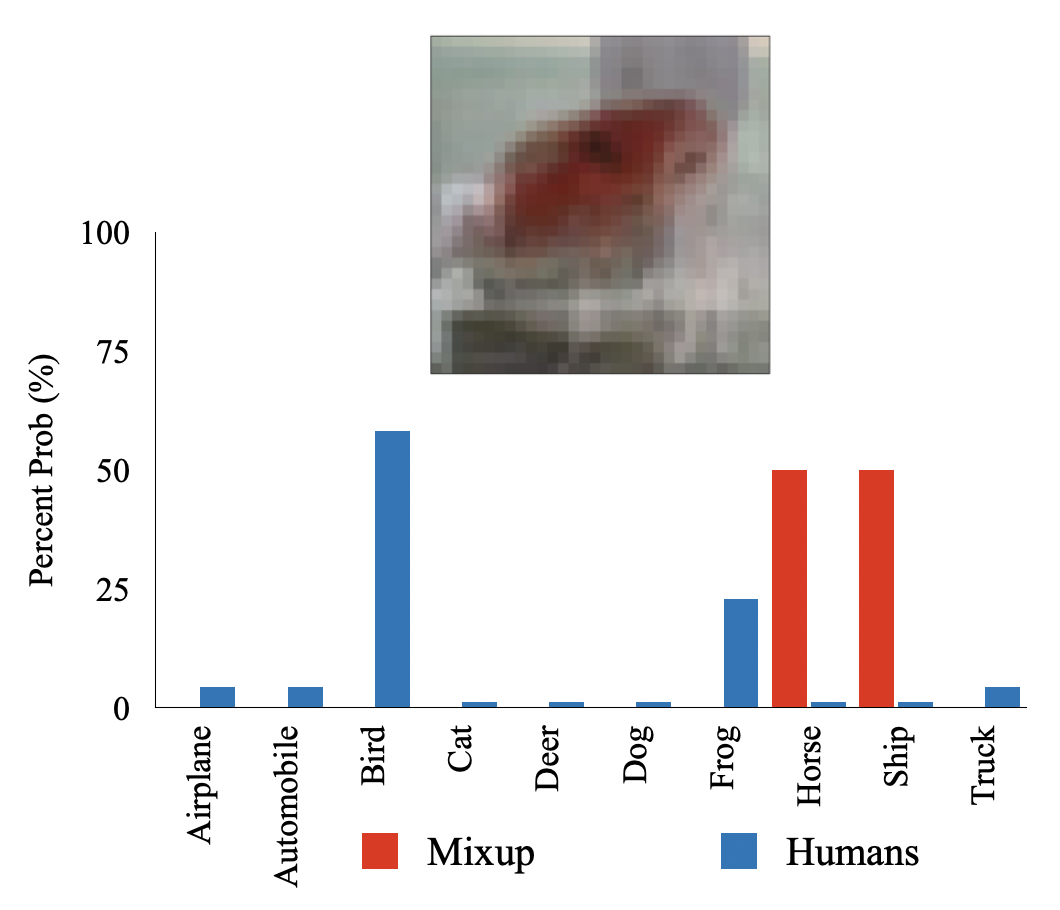}
   \caption{Example combined image ($\lambda_f = 0.5$; horse/ship) which has been relabeled by humans (blue) through the elicitation of individual soft labels our soft label elicitation (see Supplement); here, we average the individual soft labels derived from the two different participants who annotated the image. The label which would be used by \textit{mixup} is shown in red.}
    \label{fig:elicSoftLabels}
 \end{center}
 \end{figure}

We can go further in the study of human uncertainty over \textit{mixup} examples by directly eliciting soft labels from each individual over the entire space of possible classes, inspired by~\citep{selfCiteSoftLabel}. We include a preliminary investigation into eliciting richer forms of human uncertainty over \textit{mixup} examples, which lend additional nuance to the discrepancy between human perceptual judgments and the synthetic labels classically used in \textit{mixup}, in the Supplement. In particular, our primary finding is that people sometimes place probability mass on classes which are \textit{different} from the endpoint classes being combined (see Figure \ref{fig:elicSoftLabels}). We include the elicited soft labels in our release of \texttt{H-Mix}; these soft labels are small-scale at present (from $N=8$ participants, see Supplement), and we have not yet explored their computational implications, but see grounds for leveraging richer forms of human uncertainty in this vein as ripe for future work. 


\paragraph{\textit{Takeaways}} Our dataset, \texttt{H-Mix}, highlights discrepancies between humans' internal models of synthetically generated data compared to what is traditionally used in \textit{mixup}. We observe variable labeling policies on a category-pair basis and uncover a likely relationship between the strength of the mixing coefficient and ambiguity of the underlying combined images with participants' reported uncertainty in their judgments. We also preview even richer discrepancies between human percepts and the functional form of the \textit{mixup} mixing policies (see Supplement). 

\section{
Learning with Human Relabelings \textit{and Uncertainty}} 


In addressing RQ1 and RQ2, this work illuminates that human perceptual judgments do not consistently recover the parameters of the generative model traditionally used to construct data in \textit{mixup}. These findings beg the question: if we instead align synthetic examples with human perceptual judgments, how would this impact model performance? Such a question is important to consider in the pursuit of more trustworthy ML systems: better generalization, robustness, calibration, and a richer understanding of whether the models are trained on human-aligned data could all potentially engender more stakeholder trust \citep{umangTrust}. 

To that end, we consider two initial empirical studies of the impact of training on human perceptual judgments of synthetic examples: one, wherein we compare training models with varied forms of labels on the specific set of $2070$ mixed images from \texttt{H-Mix}, and another where we go beyond the collected examples and consider a first attempt at constructing a generic human-aligned label mixing policy. Here, we focus on the data collected for RQ2; i.e., for given $\tilde{x}$ how should we change $\tilde{y}$. We encourage leveraging and scaling the data collected in RQ1 for future work.  

\subsection{Relabeling Directly with \texttt{H-Mix}}

\textbf{Setup} We train a PreAct ResNet-18 \citep{resnet} and VGG-11 \citep{simonyan2014very} over 7,000 regular \texttt{CIFAR-10} images (following the split used by \citep{selfCiteSoftLabel}) combined with the 2,070 synthetically mixed images where we vary the labels. While we would ideally study human relabelings for every synthetic image that could be generated with $f$, we only have labels for a small subset and instead compare using our labels versus traditional \textit{mixup} labels over a \textit{finite, augmenting set} of combined images. 5 seeds are run per variant per architecture. Results are averaged across architectures.

\textbf{Evaluation} We evaluate a suite of metrics over 3,000 examples from \texttt{CIFAR-10H}, a dataset containing labels from many humans over the \texttt{CIFAR-10} test set \citep{peterson2019human}. We compare: cross entropy between the model-predicted and the human-derived label distributions (CE), model calibration following \citep{hendrycks2022pixmix} and robustness to the Fast Gradient Sign Method (FGSM) adversarial attack \citep{goodfellow2014explaining}, again following the set-up of \citep{selfCiteSoftLabel}.

\textbf{Leveraging Human Relabelings for ML Training} We first compare learning with our averaged human-inferred mixing parameters against the classical \textit{mixup} labels over the same 2070 synthetically-mixed images. We include sanity checks with completely random and uniform labels for the synthetic examples, as well as a baseline not including any synthetic examples (``No Aug''). Interestingly, we find in Table \ref{tab:trainAvgHuman} that aligning the mixed example labels with averaged human labels yields \textit{worse} model performance.  We think these results are worth highlighting: it is not always the case that aligning models to human perception yields performance gains, possibly due to the recently discovered U-shaped relationship between representational alignment and generalization~\citep{sucholutsky2023alignment}.

\textbf{The Value of Human Uncertainty Information} However, the human-inferred $\lambda_g$ alone does not capture the richness of human perceptual judgments over synthetic images: participants at times reported being uncertain in their inferences. Therefore, we account for human uncertainty ($\omega$) in the inference of the synthetic data generating parameter to construct softer $\tilde{y}$ (see Supplement for details). We find substantial performance boosts come from leveraging human uncertainty. Such data suggest that indeed, aligning models in accordance with human perceptual inferences could have advantages -- and suggests that confidence could offer a potent modulator signal worth considering eliciting. This is in line with core ideas from Dempster-Shafer Theory \citep{shafer1976mathematical, dempsterProbs}, that soft labels should be expressed as one set of values representing the mixture weights, and a second associated set of values representing uncertainty about that mixture (i.e., belief and plausibility).



\begin{table}[]
    \centering
        \caption{Comparing performance when varying the form of the synthetic labels on the $2070$ mixed images. Results averaged over 5 seeds, with error bars depicting 95\% confidence intervals (CIs) across seed performance.}
\begin{tabular}{llll}
\toprule
 Label Type                                      & CE             & FGSM       & Calib   \\
\midrule
 Regular                         &&&      \\
 (No Aug)                                & 2.02$\pm$0.12 & 13.12$\pm$2.65 & 0.28$\pm$0.011      \\
 + Random                                   & 2.11$\pm$0.13 & 12.81$\pm$2.84  & 0.24$\pm$0.014      \\
 + Uniform                                  & 2.16$\pm$0.14 & 12.71$\pm$2.79  & 0.25$\pm$0.012      \\
 + \textit{mixup}                                    & 1.65$\pm$0.11 & 10.62$\pm$2.44 & 0.23$\pm$0.005   \\
 + Ours                         &&&      \\
 (Relabel)                          & 1.78$\pm$0.12 & 11.69$\pm$2.90 & 0.24$\pm$0.009      \\
(Relabel \& $\omega$)      & \textbf{1.48$\pm$0.06} & \textbf{8.89$\pm$1.59}  & \textbf{0.19$\pm$0.001}      \\
\bottomrule
\end{tabular}
    \label{tab:trainAvgHuman}
\end{table}

\subsection{Generalizing Relabeling}

So far, we have focused on varying the labels of a pre-supposed augmenting set of mixed images; however, the set was comparatively small (2070 images) and therefore does not directly mimic the \textit{mixup} learning paradigm. In practice, \textit{mixup} is typically applied over the entire dataset; that is, on each batch, a new mixing coefficient is sampled, resulting in often entirely new images being generated per batch. It is infeasible to consider recruiting human participants to relabel every such image. Automated human-aligned labeling policies are therefore worth considering. We argue that our data offers a prime starting point to explore such questions.

We offer a preliminary alternative label mixing policy based on the human data we have collected in \texttt{H-Mix}. Inspired by the non-linearities we observe at a category level, we use \texttt{scipy.curve\_fit} to fit a logistic function per category pair. For each batch, we swap in our label mixing policy to map from the sampled generating mixing coefficient to an approximately more human-perceptually aligned coefficient. Such fits only account for humans' relabelings, not their uncertainty. Accounting for human confidence in automated label policies is a ripe direction for future work. 

\textbf{Setup} We follow the same ensembling and evaluation methodology laid out in Section 5.1, but now run traditional \textit{mixup} following \citep{mixup} where generating mixing coefficients are sampled from a $Beta(1,1)$ distribution (i.e., uniform on $(0,1)$). 

\textbf{Results} We observe (see Table \ref{tab:autoLabel}) a striking parity in performance across models. These data highlight that with the addition of even a relatively small number of human annotations through \texttt{HMix} to alter the labeling policy, we find that robustness to adversarial attacks increases at negligble cost to performance or calibration. As in \citep{sucholutsky2023alignment}, human representation alignment may be useful for other downstream, untested tasks: training on more human-aligned data-generating policies could induce functional fits that are preferable to stakeholders even if we see no objective improvement along particular performance measures. We recommend such studies for future work. 



\begin{table}[]
    \centering
        \caption{Training with mixing policies fitted per category pair, compared against full \textit{mixup}. Results averaged over 5 seeds, with 95\% CI error bars.}
\begin{tabular}{llll}
\toprule
 Label Policy   & CE             & FGSM      & Calib   \\
\midrule
 \textit{mixup}           & \textbf{1.15$\pm$0.08} & 7.46$\pm$2.40  & \textbf{0.10$\pm$0.01}      \\
  Human-Fits (Ours) & 1.16$\pm$0.08 & \textbf{7.32$\pm$2.27} & \textbf{0.10$\pm$0.01}      \\
\bottomrule
\end{tabular}

    \label{tab:autoLabel}
\end{table}

\textbf{\textit{Takeaways}} Human perceptual judgments can be leveraged to construct alternative synthetic data-generating policies to train ML systems; however, such induced methods of aligning with (approximations) of human perception are not automatic salves. Our results highlight that constructing more human-aligned label policies, particularly through capturing and representing human uncertainty, is promising, but more work is needed before generalizing conclusions.

\section{Discussion}
\textbf{(Mis)alignment of Mixup Examples} 
Through a series of novel user studies, we uncover that the synthetic examples used in \textit{mixup} do not consistently align with humans' perceptual representations. We find indications that participants' \textit{uncertainty} in their inferred mixing coefficients tracks with the degree of ambiguity of the original images that are combined. As we have begun to explore empirically, such relabeling may impact downstream model performance: re-aligning mixup labels with humans' reported judgments can impact learning, with human uncertainty seemingly poised to provide a strong supervisory signal -- corroborating \citep{peterson2019human, collins2022structured, iliaSupervision}. The collation of humans' inferences of the \textit{mixup} generative parameters could also be used to benchmark whether models are aligned with human percepts, say if \texttt{H-Mix} is used as a held-out or probe set \citep{gruber2018perceptual}. We recommend such directions for future work, particularly those focused on the uncertainty elicitation in \texttt{H-Mix}. We release additional soft labels over mixed examples which further highlight human perceptual misalignment (see Supplement). 

\textbf{Scaling Human-Centric Data Relabeling} A key challenge for human-centric relabeling of synthetically-generated data (not unique to \textit{mixup}) is that a nearly infinite variety can be generated. It is not reasonable to expect humans to 
judge \textit{all} possibilities, nor to provide their uncertainty over all labels.
Any attempt at human-in-the-loop relabeling faces the obstacle of identifying which examples to relabel, and how to handle cases that cannot be relabeled. While we take steps to address the latter through fitting generic functions per class pair that enable sampling of arbitrary mixing coefficients, we highly encourage researchers to consider leveraging our \texttt{H-Mix} to develop alternative human-grounded automated synthetic data policies.


To address the former, we encourage looking to smarter ways to select examples to query people over -- rather than random selection as we have done -- such as~\citep{Liu2021LAST,liu2017iterative}. Additionally, our results raise the related question: are there particular relabelings that are \textit{hurting} model performance? Prior works have demonstrated how cleaning data can reduce model error \citep{pleiss2020identifyingMislabeled}. We encourage future work in this direction in the context of \texttt{H-Mix}. 
Additionally, our results raise the related question: are there particular relabelings that are \textit{hurting} model performance? Prior works have demonstrated how cleaning data can reduce model error \citep{pleiss2020identifyingMislabeled}. We encourage future work in this direction in the context of \texttt{H-Mix}.

\textbf{Limitations} Thus far, we only consider human validation and relabeling of \textit{mixup} labels for a single image classification dataset, \texttt{CIFAR-10}. This dataset is 
low-resolution. Thus, the endpoint images -- and the combinations of images -- can be ambiguous and challenging to interpret. It is possible that we may find humans to be more, or less, aligned with the generative parameters for different image datasets, 
or for entirely different data modalities, e.g., audio or video. We encourage the application of the \texttt{HILL MixE Suite} paradigm to other datasets. Moreover, as we have many category pairs -- arising even from just 10 categories -- we do not have a substantial number of synthetic examples \textit{per} category pair (i.e., 46 synthetically-mixed images for each of the 45 category pairs). This could impact the stability of the category boundaries we elicit, e.g., potentially leading to breaks of monotonicity (see Supplement A). Further, as with many web-based human elicitation studies, it is not always clear whether the responses returned arise from individual differences in perception, participant noise, or malicious behavior \citep{lease2011quality, gadiraju2015understandingCrowdsource}. We also do not train participants to provide calibrated uncertainty; uncertainty judgments included in \texttt{H-Mix} -- while empirically useful for training -- could be infused with classical biases in humans' probabilistic self-reports \citep{lichtenstein1977calibration, kahneman1996reality,uncertainJudgments, sharot2011optimism}. We also highlight that, aside from repeat trials, we are unable to capture whether participants' percepts fluctuate -- such instability is certainly a possibility when considering cognitive neuroscience research around perceptual dominance \citep{blake2002visual}. 

\textbf{Extending to New Synthetic Data Paradigms}
In this work, we focused on the synthetic data classically used in \textit{mixup}, as the simplicity of the data generating process -- a single mixing coefficient parameter -- enables us to precisely compare human versus traditional parameterizations of the synthetic data construction process. We hope our work spurs further study of aligning synthetic data generation with human perception and motivates the design of more human-aligned synthetic data to improve ML systems, particularly those focused on the interplay between model and human \textit{uncertainty}. We release the code of all interfaces included in our \texttt{HILL MixE Suite}, which we hope will empower researchers with additional tools to investigate humans' percepts over synthetically-constructed data. For instance, our \texttt{Select-Shuffled} interface could readily be extended to elicit stakeholders' preferences, in the form of selection, over any collection of constructed synthetic examples. As demonstrated in \citep{instructGPT}, scalable human preference elicitation has wide utility. 

\section{Conclusion}

Through a series of human participant elicitation studies, we find that the synthetic examples generated via \textit{mixup} differ in fundamental ways from human perception, suggesting misalignment of the data and label mixing policies. We offer early indications that collating humans' percepts of these synthetic examples could impact model performance, particularly when modulated \textit{by elicited human uncertainty}. Our work further motivates the design of automated relabeling procedures for synthetic examples which leverage elicited human data (e.g., training a model to predict a likely human's mixing coefficient) to sidestep inherent issues with scaling human annotation over the space of possible synthetic examples, particularly in eliciting and utilizing human uncertainty. Synthetic data of all kinds are proliferating: we encourage more researchers to consider these data from a human-centric perspective; i.e., investigating whether the samples align with human percepts, and if not, whether altering labels -- specifically via human uncertainty -- can yield safer, more reliable models with improved generalization.

\section*{Acknowledgments}
We thank (alphabetically) Federico Barbero, Alan Clark, Krishnamurthy (Dj) Dvijotham, Benedict King, Tuan Anh Le, Haitz Sáez de Ocáriz Borde, N. Siddharth, Joshua Tenenbaum, Richard E. Turner, Vishaal Udandarao, and Lio Wong for helpful discussions. We also thank our participants on Prolific, and our reviewers for incredibly helpful feedback.

KMC is supported by a Marshall Scholarship. UB acknowledges support from a JP Morgan AI PhD Fellowship and from DeepMind and the Leverhulme Trust via the Leverhulme Centre for the Future of Intelligence (CFI). BL acknowledges support under the ESRC grant ES/W007347, and AW acknowledges support from a Turing AI Fellowship under grant EP/V025279/1, The Alan Turing Institute, and the Leverhulme Trust via CFI. IS is supported by an NSERC fellowship
(567554-2022). 

\bibliography{collins_256}

\onecolumn 

\section*{Related Work}

Our work connects most closely to human-in-the-loop data augmentation and the expansive literature surrounding human categorical perception from the cognitive science community, as well as ongoing efforts in the machine learning community to develop more efficacious \textit{mixup}-based data and label mixing functions. 

\subsection*{Human-in-the-Loop Data Augmentation}
Incorporating expert feedback into the learning procedure has received increasing attention~\citep{chen2022perspectives}.
In particular, previous work has considered incorporating humans ``in the loop'' for data augmentation. For instance, DatasetGAN~\citep{datasetGAN} employs human participants to label GAN-generated images and feeds these back to the model to generate more synthetic data. \citep{counterfactuallyAugmentedHuman} similarly incorporate human feedback by having humans \textit{create} counterfactual samples, and has been shown to be an efficient method to adjust model behavior \citep{efficacyCounterfactual}. Other works have considered employing humans to provide ``rationales'' about examples to improve data-efficiency and downstream modeling performance \citep{zaidan2007using}. Here, we marry these ideas in the context of \textit{mixup} by eliciting data and label-mixing function parameters to align with human percepts.  

\subsection*{Human Categorical Perception} In cognitive science, eliciting humans' judgments over synthetically-constructed examples is a tried-and-true method to characterize human category boundaries \citep{newell2002categorical,folstein2013category,feldmanCategoricalPerception, folsteinFactorizedvBlended}. Such studies often reveal a non-linear structure of humans' percepts. For instance, in the audio domain, the identification of vowel categories has been found to demonstrate ``warping'' close to prototypical category members -- known as the ``perceptual magnet effect'' \citep{perceptualMagnetEffect, perceptualMagnetStats}. Similar nonlinearities have been found in the perception of boundaries between face identities \citep{beale1995categorical} and the transitions between 3D shapes \citep{newell2002categorical, morphSpacesShape}. Our linearly interpolated stimuli are similar in spirit to the morphological trajectories used in these works, as well as other synthetically-combined images \citep{hybridImages}. \citep{gruber2018perceptual} also consider 50/50 mixed images; however, their elicitation involves open-ended judgments which does not permit the same kind of data and label mixing alignment studies as our methods more directly elicit human-inferred generative parameters. Our work also connects to other non-linear perceptual phenomena encountered in the visual domain; namely, binocular rivalry, whereby present participants with a different image in each eye has been shown to induce oscillatory percepts \citep{blake2002visual, tong2006neural}. 



\subsection*{Other \textit{mixup}-Based Synthetic Data Schemes} Many alternative \textit{mixup} data and label mixing functions have been proposed \citep{manifoldMixUp, cutMix, puzzleMix, coMixup, hendrycks2022pixmix}. Closest to our work, \citep{sohn2022genlabel} highlight particular issues with the linear interpolation in label space on the learned topology of the model's category boundaries and instead utilize a Gaussian Mixture Model (GMM)-based relabeling scheme to construct ``better'' labels than those used in baseline \textit{mixup}. Additional work on learning better pseudo-labels over \textit{mixup} samples have been proposed \citep{pseudoLabelMixup, cascante2020curriculum, fixmatch, Qiu2022DHT}. Similarly, Between-class (BC) learning \citep{bcLearningAudio,bcLearning} proposes hand-crafted adjustments to label construction to better align with human perception based on waveform modulations; however, to our knowledge, no previous works have \textit{directly} considered incorporating humans in-the-loop for either the construction of \textit{mixup} samples or associated relabeling.

\section*{Additional Notes on \texttt{H-Mix}}

\subsection*{Human Subject Experiments}

We include additional details on our human elicitation studies. For all experiments, we require participants speak English as a first-language and reside in the United States. Across all experiments, the mean age for participants was 37.5 years old ($\pm$ standard deviation of 12.7 yrs) . The self-reported sex breakdown was approximately 57\% male and 43\% female.

\paragraph{Elicitation (RQ1)} Each participant sees a total of 32 mixed images, where the final two are repeats. Repeats are primarily used here to measure raters' internal consistency\footnote{Participants' selections, for each interface type, change by a median of $0.1$ in repeat trials, suggesting some inconsistencies in participants' judgments which persists across elicitation method.}. The median time taken per participant per image as 9.30 and 11.01 seconds for the \texttt{Construct} and \texttt{Select-Shuffled} interfaces, respectively. A bonus was offered to encourage participants to provide responses which would match what other participants would provide; we applied this bonus to all participants post-hoc resulting in the average participant being paid at a rate of \$11.78. 

\paragraph{Multiple Interface Styles (RQ1)} Why do we consider two styles of elicitation interfaces? We reason that the first interface could be prone to ordering effects -- an astute participant could realize that they can count out where the midpoint is located. This led us to design the second variety (\texttt{Select-Shuffled}) wherein the participant sees all images shuffled simultaneously. We hypothesize that \texttt{Construct} could induce responses biased by the participant's starting position. To probe this, we run two sub-variants wherein participants start from either $\lambda_f = 0.1$ or $\lambda_f = 0.9$. 

\paragraph{Elicitation (RQ2)} Each participant sees $59-62$ images, where two images are repeated. Repeats are placed at the end and correspond to the images presented on trials 15 and 20, respectively\footnote{We observe a median difference of $0.03$ and $0.05$ in the inferred mixing coefficient and confidence on repeat trials, indicative of high intra-annotator consistency.}. The order of the images presented in a batch, as well as the order of the endpoint labels displayed for a given image, are shuffled across participants. We follow the same third-person perspective prompting in Section 3 from \citep{efficientElic}. Participants are asked ``what combinations of classes'' they thought other participants would say is ``used to make'' each image, and ``how confident'' they thought other participants would be in their estimate. Responses are indicated on a slider per question. An example survey screen can be seen in Fig. \ref{fig:inferCoeffInterface}. Subjects took a median of 8.41 seconds per image and were payed at a rate of \$8/hr, with an optional bonus which sought to encourage participants to provide calibrated confidence estimates, similar to that of \citep{vodrahalli2021humans}; the bonus was applied to all participants post-hoc. Each mixed image was seen by at least two different participants each. Our interface is depicted in Fig. \ref{fig:inferCoeffInterface}. 



\subsection*{Break from Monotonicty}

\begin{figure}[h!]
 \begin{center}
 \includegraphics[width=0.8\linewidth]{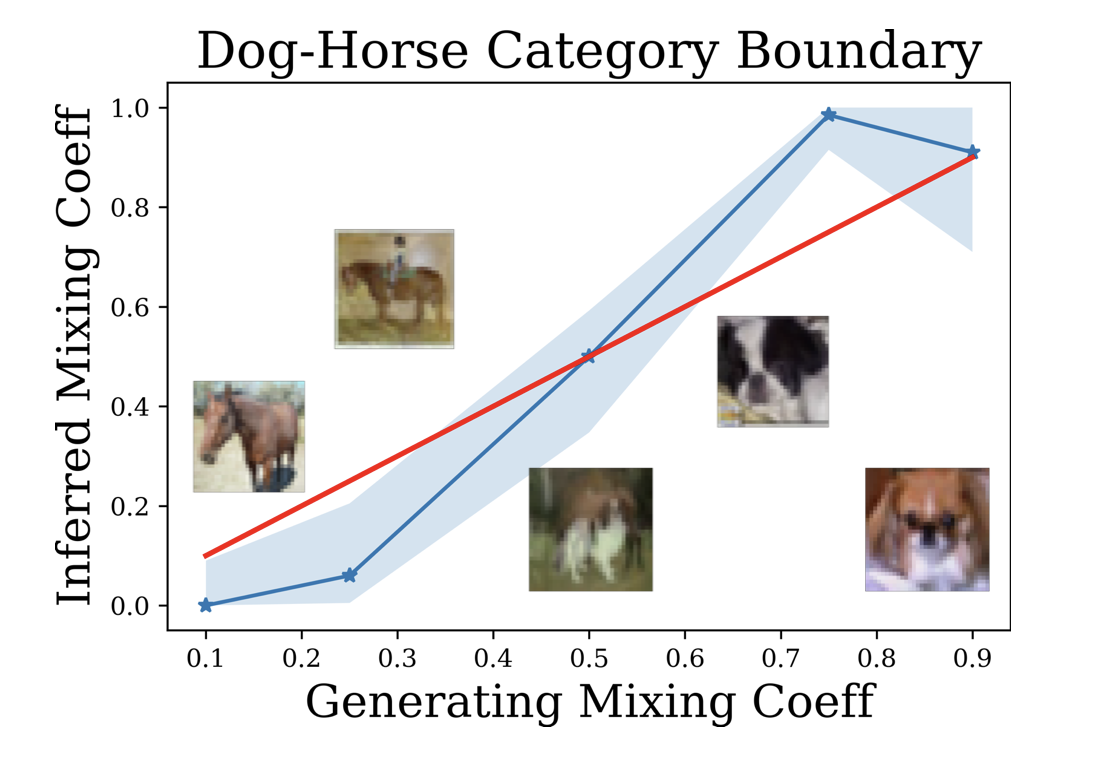}
   \caption{Category boundary elicited from human participants involves a break with monotonicity.}
 \label{fig:nonMonotonic}
 \end{center}
 \end{figure}

For users of \texttt{H-Mix}, it is worth noting that we do encounter some breaks with monotonicity (see Fig. \ref{fig:nonMonotonic}) in a few of the aggregated ``category boundaries.'' We reason this could be in part due to several aspects of our set-up. First, our study involved irregular sampling across the space of mixing coefficients we consider: the 50/50 point is enriched. We ran two phases of elicitation: in the first, we sampled $6$ image classes per pair to be shown for three mixing coefficients: 0.5, and one chosen randomly from each of the sets \{0.1, 0.25\} and \{0.75, 0.9\}, respectively (810 images of the 2070). All 1260 other images are shown for a single mixing coefficient sampled uniformly from the set. Second, while we have human judgments for over 2000 total images, there are less than 50 synthetic images considered for each category pair, giving any participant noise -- or the odd image  -- greater leverage to impact trends. We encourage others to use \texttt{HILL-MixE Suite} and continue to scale this work and elucidate the stability of the inferred mixing coefficient category boundaries we begin to hint at here. 

\section*{Confidence-Based Smoothing Details}

We include further details of our methodology for leveraging human-provided confidence to construct $\tilde{y}$ introduced in Section 5. Human-derived soft labels have been demonstrated to be valuable for learning \citep{softlossMed, peterson2019human, selfCiteSoftLabel, sandersambiguous}. We transform humans' reported confidence into a smoothing parameter to induce softness using an exponentially-decaying function of human-provided confidence $\omega$: $a * (b^\omega)$; here, $a = 50, b = 0.0001$. We use the transformed confidence for additive smoothing on the two-category $\tilde{y}$, spread mass accordingly across the full gamut of classes. That is, we use smooth the mass between a completely uniform distribution and a ``two-hot'' label which uses the human-derived relabeling. Parameters $a, b$ are selected using a held-out set of regular \texttt{CIFAR-10} images (from $a \in \{5, 10, 15, 25, 50, 100\}, b \in \{0.00001, 0.0001, 0.001, 0.01, 0.1\}$). We recommend the consideration of alternate smoothing functions, which could, for instance, account for miscalibration in humans' reported confidence.  

Further, we compare the impact of learning with aggregated versus de-aggregated participants' predictions. In Section 5, we considered learning with relabelings averaged across participants for a mixed image, and smoothed with confidence reports averaged across participants. Here, we consider instead separating out participants' responses to learn with individual relabelings smoothed by individual confidence, closely related to \citep{wei2022aggregate}. We find in Table \ref{tab:trainHumanConf} that learning with \textit{de-aggregated} data could potentially offer greater performance gains. However, as \citep{wei2022aggregate} discuss: whether to aggregate can depend on many factors. Our empirical findings support the need for tailoring label construction in context. 

\begin{table*}[]
    \centering
        \caption{Varying whether to aggregate when using incorporating human confidence $\omega$ in label construction.}
\begin{tabular}{llll}
\toprule
 Label Type                                      & CE             & FGSM       & Calib   \\
\midrule
Ours (Avg with $\omega$)      & 1.48$\pm$0.06 & 8.89$\pm$1.59  & \textbf{0.19$\pm$0.01}      \\
 Ours (Separated with $\omega$) & \textbf{1.44$\pm$0.11}  & \textbf{8.33$\pm$1.92}  & \textbf{0.19$\pm$0.01}      \\
\bottomrule
\end{tabular}

    \label{tab:trainHumanConf}
\end{table*}

\section*{Interfaces Included in \texttt{HILL MixE Suite}}

\begin{figure}[h!]
 \begin{center}
 \includegraphics[width=0.95\linewidth]{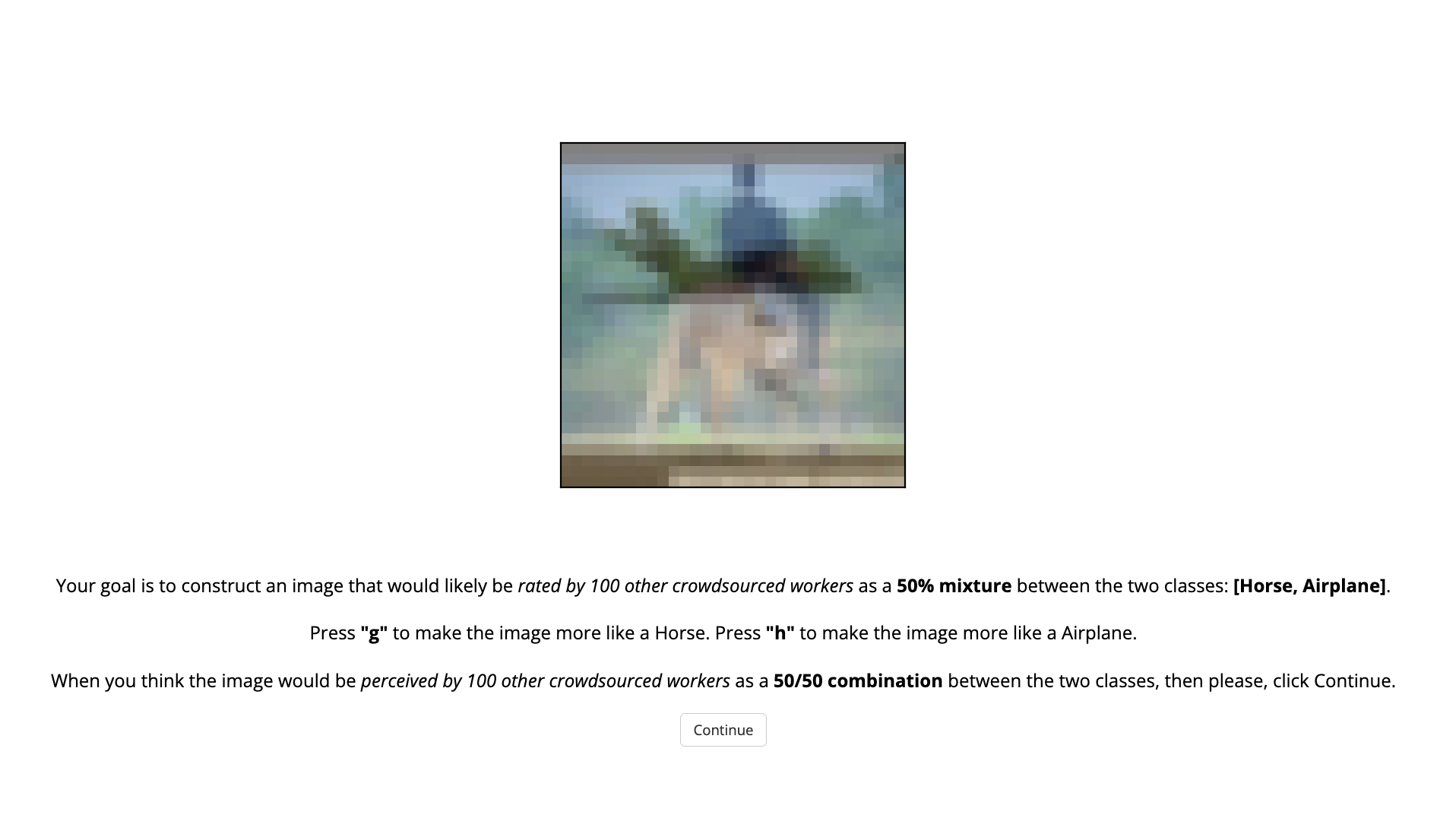}
  \caption{Construct interface where participants press arrow keys to select $\tilde{x}.$}
    \label{fig:clickNextInterface}
 \end{center}
 \end{figure}

\begin{figure}[h!]
 \begin{center}
 \includegraphics[width=0.95\linewidth]{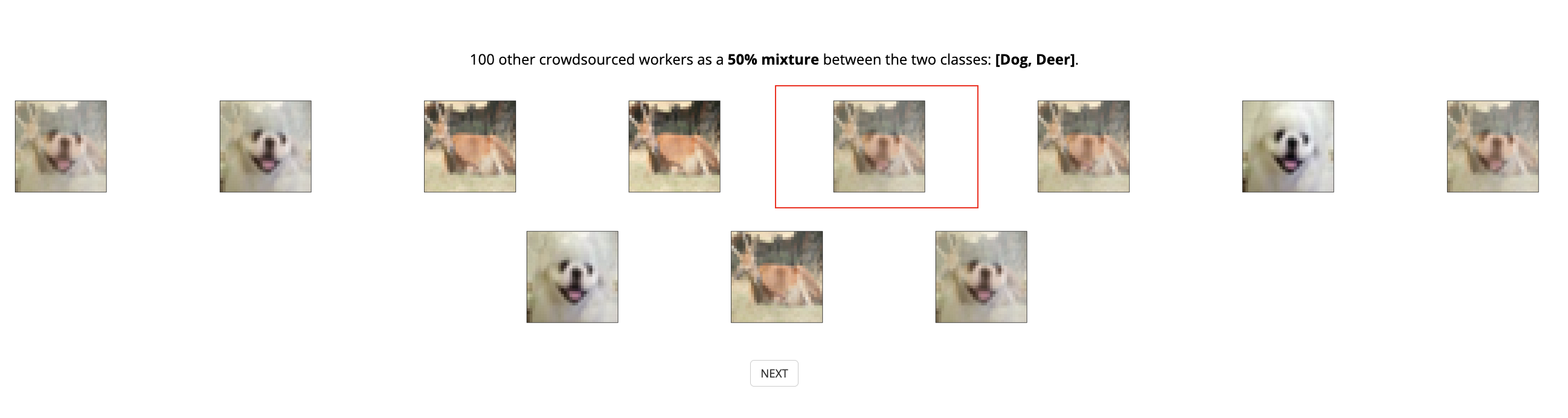}
  \caption{Interface for the selection of a given $\lambda_g$ from a set of possible mixed images.}
    \label{fig:synthDataSelectInterface}
 \end{center}
 \end{figure}

\begin{figure}[h!]
 \begin{center}
 \includegraphics[width=0.95\linewidth]{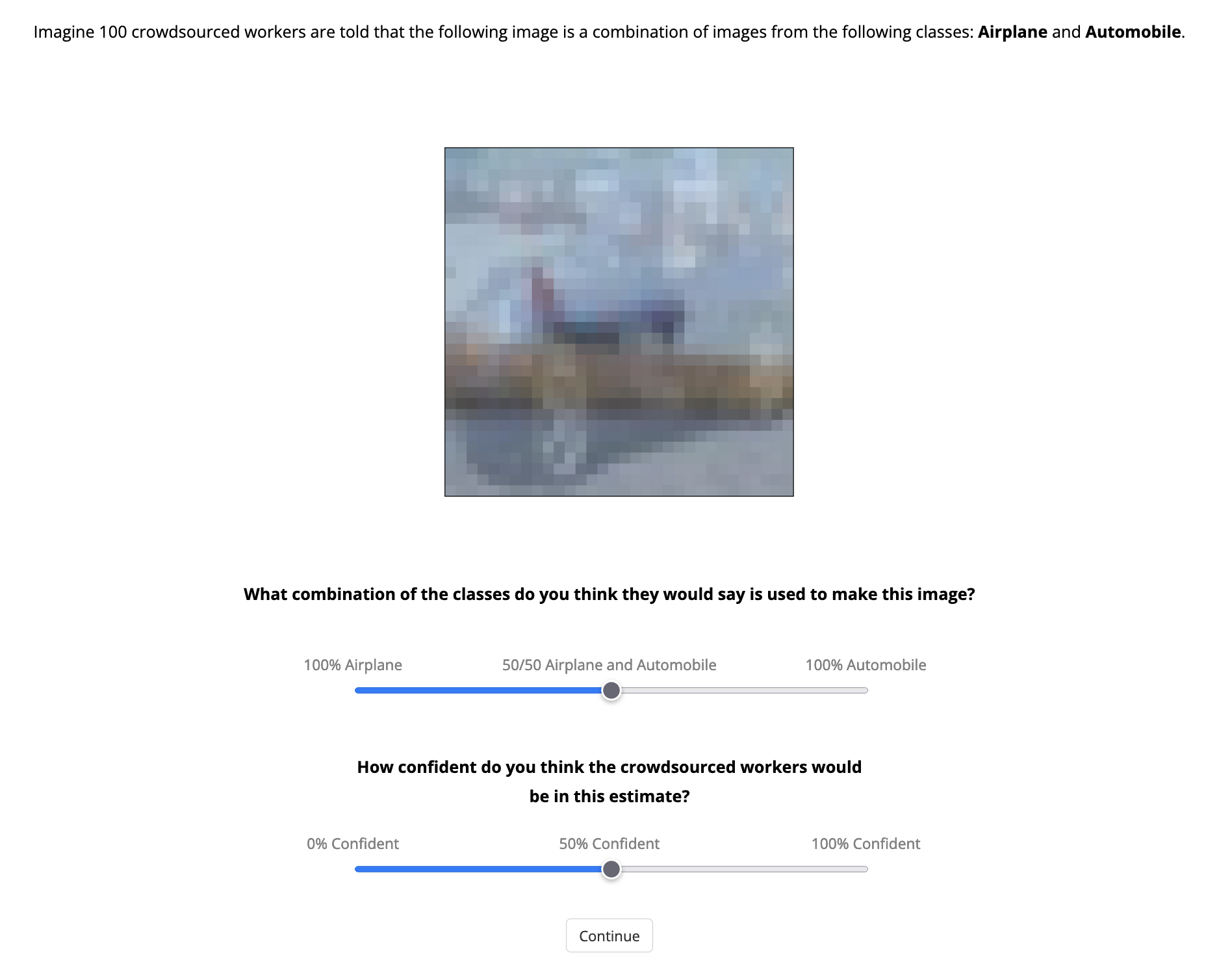}
   \caption{Interface for inferring the \textit{mixup} generative label parameter and providing confidence in such inference.}
    \label{fig:inferCoeffInterface}
 \end{center}
 \end{figure}

We display sample pages of the interfaces created and used in this work, which we release as part of \texttt{HILL MixE Suite}. Interfaces for Section 3 are shown in Figs. \ref{fig:clickNextInterface} and \ref{fig:synthDataSelectInterface}; the interface used Sections 4 is depicted in Fig. \ref{fig:inferCoeffInterface}. 

\section*{Alternative Synthetic Example Category Composition Elicitation} 

\begin{figure*}[h!]
 \begin{center}
 \includegraphics[width=0.98\linewidth]{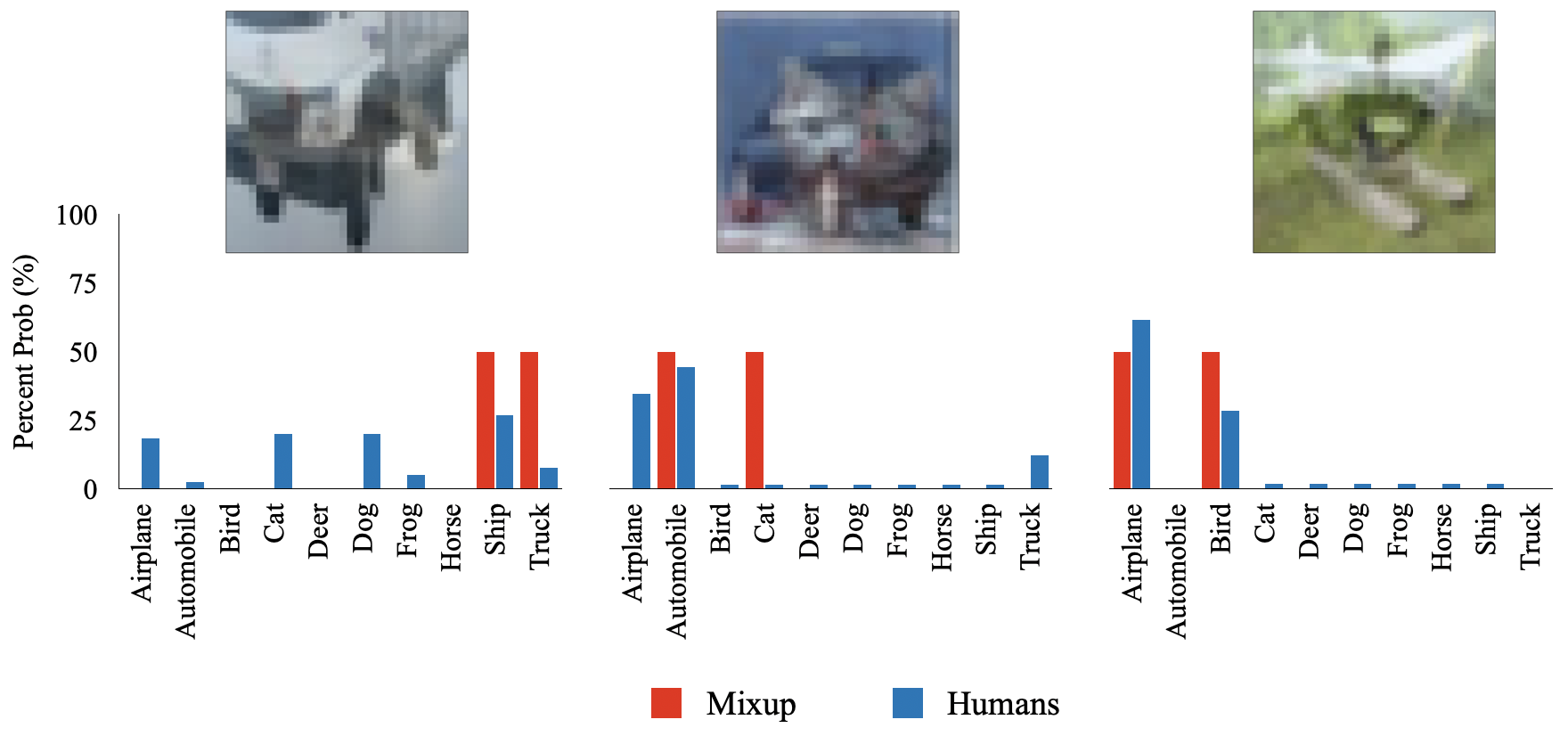}
   \caption{Additional example soft labels elicited from individuals. Original \textit{mixup} label for each associated image is shown in red; the soft label elicited from humans (averaged over two individuals) is shown in blue. The left and center examples involve substantial discrepancies between human percepts and the label which would be used in \textit{mixup}; the rightmost image highlights that some percepts do match the underlying mixing components (even without being informed of the underlying classes). Examples are deliberately chosen to illustrate the range of soft labels elicited; all examples are include in \texttt{H-Mix}.}
    \label{fig:elicSoftLabelsMore}
 \end{center}
 \end{figure*}

Given human participants are uncertain about the underlying mixing coefficient in a number of cases, we consider whether the category composition typically used in \textit{mixup} -- e.g., placing mass only on the labels of the images used to form the synthetic combined sample -- are reasonable. As demonstrated in the main text and in Fig. \ref{fig:elicSoftLabelsMore}, a synthetic \textit{mixup} image may look like something else entirely. 

We therefore consider a follow-up small-scale human elicitation study wherein we relax the \textit{mixup} assumption that the label mixing function must output a label constructed only from the two classes used to form the mixed image -- and instead collect $\tilde{y}$ \textit{directly} by showing the mixed image 
to human annotators in the form of soft labels. This provides a comparison to the previous human-annotated endpoint label mixing coefficients, and can further inspire useful designs for the label mixing policy. 

\subsection*{Study Design} We recruit $N=8$ participants again from Prolific \citep{palan2018prolific}, yielding soft labels over a total of $100$ mixed images. Each participant saw 25 mixed images; each mixed image of the $100$ was seen by two participants. The images are drawn from the same set of stimuli created in Section 4; however, here, we only show images with a mixing coefficient $\in \{0.25, 0.5, 0.75 \}$. Participants are told that images are formed by combining other images, and are asked to provide what they think others would see in the image. Participants are asked to specify what others would view as the most probable category with an associated percentage (on a scale of 0-100), an optional second most probable category with a probability, and any categories that would be perceived as definitely not in the image. Again employing the third-person viewpoint framing borrowed from \citep{efficientElic}. We rely on the soft label elicitation interface proposed in~\citep{selfCiteSoftLabel} and modify the instructions to be better suited combinations of images. Following \citeauthor{selfCiteSoftLabel}, we construct ``Top 2 Clamp'' labels with a redistribution factor of 0.1, which controls how we spread mass over any categories still leftover as ``possible'' once accounting for those ruled out as definitely not possible. 


\subsection*{Analyzing Elicited Soft Labels for Synthetic Images}

We explore the correspondence between the elicited category compositions of the mixed images with the labels that would be used to generate the mixed image (as would be used in traditional \textit{mixup}; i.e., placing mass only on two categories). While participants did tend to place probability mass on the generating endpoints that correlated with the mixing coefficient used (Pearson $r = 0.52$), interestingly, we find that participants report thinking that 38.3\% ($\pm$0.6\%) of the probability mass of a label should be placed on \textit{different} classes from those which are used to create the image. This is remarkable and suggests that mixed images \textit{do not} consistently look like the labels used to create them, corroborate similar trends found in \citep{gruber2018perceptual} wherein humans endorse categories which are not present in the image. Hence, alternative labelings even beyond the kind we explore in the main text may be preferred which are more aligned with human percepts. Examples of such labeled mixed images are shown in Fig.~\ref{fig:elicSoftLabelsMore} and the main text. 

\paragraph{\textit{Takeaways}} The typical two-category labels used in \textit{mixup} do \textit{not} consistently match human perception. 
We find that human annotators often assign probabilities to alternate classes when asked to label a mixed image. This suggests that the pursuit of aligning synthetic data labeling to match human perception, at least for the synthetic data constructor used in \textit{mixup}, warrants the design of alternative label mixing functions $g_\text{rich}$ which yield richer label distributions over a broader range of categories.

\end{document}